\documentclass[letterpaper, 10 pt, conference]{ieeeconf}  
\IEEEoverridecommandlockouts                              
\makeatletter
\newif\if@restonecol
\makeatother

\usepackage{cite}
\usepackage{amsmath,amssymb,amsfonts}
\usepackage{algorithmic}
\usepackage{graphicx}
\usepackage{textcomp}
\usepackage{xcolor}
\usepackage{booktabs}
\usepackage{bbding}
\usepackage{tikz}
\usepackage{bm}
\usepackage[ruled,linesnumbered,noline]{algorithm2e}
\title{\LARGE \bf
RhoMorph: Rhombus-shaped Deformable Modular Robots for Stable, Medium-Independent Reconfiguration Motion
}

\author{Jie Gu$^{1}$, Yirui Sun$^{1}$, Zhihao Xia$^{1}$, Tin Lun Lam$^{2,3}$, Chunxu Tian$^{1,\dag}$, Dan Zhang$^{1,4,\dag}$ % <-this % stops a space
\thanks{*This work was supported by the National Nature Science Foundation of China (grants 52305012).}% <-this % stops a space
\thanks{$^{1}$Institute of AI and Robotics, Academy for Engineering \& Technology, Fudan University, Shanghai 200433, China.}
\thanks{$^{2}$School of Science and Engineering, The Chinese University of Hong Kong, Shenzhen, China.}
\thanks{$^{3}$Shenzhen Institute of Artificial Intelligence and Robotics for Society.}
\thanks{$^{4}$Department of Mechanical Engineering, The Hong Kong Polytechnic University, Hung Hom.}%
\thanks{$^{\dag}$Corresponding author is Chunxu Tian, email: chxtian@fudan.edu.cn and Dan Zhang, email: dan.zhang@polyu.edu.hk}
}

\begin{document}

\maketitle
\newcommand{\SLG}{\text{SLG}}
\newcommand{\AB}{\text{AB}}
\newcommand{\AC}{\text{AC}}
\newcommand{\AD}{\text{AD}}
\newcommand{\BD}{\text{BD}}
\newcommand{\BC}{\text{BC}}
\newcommand{\CD}{\text{CD}}
\newcommand{\BA}{\text{BA}}
\newcommand{\CA}{\text{CA}}
\newcommand{\DA}{\text{DA}}

\newcommand{\Outer}{\text{Outer}}
\newcommand{\Inner}{\text{Inner}}
\newcommand{\Middle}{\text{Middle}}
\newcommand{\A}{\text{A}}
\newcommand{\B}{\text{B}}
\newcommand{\C}{\text{C}}
\newcommand{\D}{\text{D}}
\newcommand{\dk}{\text{dk}}
\thispagestyle{empty}
\pagestyle{empty}

%%%%%%%%%%%%%%%%%%%%%%%%%%%%%%%%%%%%%%%%%%%%%%%%%%%%%%%%%%%%%%%%%%%%%%%%%%%%%%%%
%%%%%%%%%%%%%%%%%%%%%%%%%%%%%%%%%%%%%%%%%%%%%%%%%%%%%%%%%%%%%%%%%%%%%%%%%%%%%%%%
\begin{abstract}
%等待替换
% In this paper, we presented a deformable planar lattice modular self-reconfigurable robot (MSRR), called RhoMorph, shaped as rhombus. the module 主要由平行四边形骨架以及one actuator 布置在中心 to generate 沿着对角线的折展。我们的设计初衷是以极简的控制使单个模块产生应有的功能诸如morphing，connection，locomotion来辅助完成模块之间的重构运动，这种重构是不需要依赖周围介质的，且能够连续的稳定变换的。这样的能力可以赋予机器人在任何环境包括海陆空，外太空，能够自由的变形产生不同的构型。针对RhoMorph的结构特点，我提出一种名为morphpivoting的基元运动区别于其他msrr的运动形式，并为之提出了一种连续执行的策略。最后，一系列的实验验证了模块的稳定的重构能力，以及模块的各项精度。
In this paper, we present RhoMorph, a novel deformable planar lattice modular self-reconfigurable robot (MSRR) with a rhombus shaped module. Each module consists of a parallelogram skeleton with a single centrally mounted actuator that enables folding and unfolding along its diagonal. The core design philosophy is to achieve essential MSRR functionalities such as morphing, docking, and locomotion with minimal control complexity. This enables a continuous and stable reconfiguration process that is independent of the surrounding medium, allowing the system to reliably form various configurations in diverse environments. To leverage the unique kinematics of RhoMorph, we introduce morphpivoting, a novel motion primitive for reconfiguration that differs from advanced MSRR systems, and propose a strategy for its continuous execution. Finally, a series of physical experiments validate the module's stable reconfiguration ability, as well as its positional and docking accuracy.

\end{abstract}

%%%%%%%%%%%%%%%%%%%%%%%%%%%%%%%%%%%%%%%%%%%%%%%%%%%%%%%%%%%%%%%%%%%%%%%%%%%%%%%%
%%%%%%%%%%%%%%%%%%%%%%%%%%%%%%%%%%%%%%%%%%%%%%%%%%%%%%%%%%%%%%%%%%%%%%%%%%%%%%%%
% section 1
\section{Introduction}

Traditional Modular Self-Reconfigurable Robots (MSRRs) are able to deploying to versatile configurations to adapt to different environment and reusable across a wide range of missions, often built with rigid modules in chain, lattice, mobile, hybrid, or freeform form\cite{liang2020freebot,tu2022freesn,belke2017mori,spinos2017vtt,castano2000conro,swissler2018fireant,zhao2024snail-nature,zheng2025Rs-modcubes}. 

With the continuous emergence of new modular robot designs, module structures are no longer confined to a single, well-defined configuration but instead exhibit multiple overlapping attributes. A detailed review of such overlapping and fuzzy classifications is provided in \cite{liang2023decoding}. Different attributes offer advantages for different functional objectives; for example, modules with mobile attributes enhance locomotion capabilities\cite{liu2023smores}, those with chain attributes are better suited for manipulation tasks\cite{sprowitz2014roombots}, and those with lattice attributes facilitate the rapid construction of temporary structures\cite{zhao2023starblocks}.

\begin{figure}[t]
\centerline{\includegraphics[width=\linewidth]{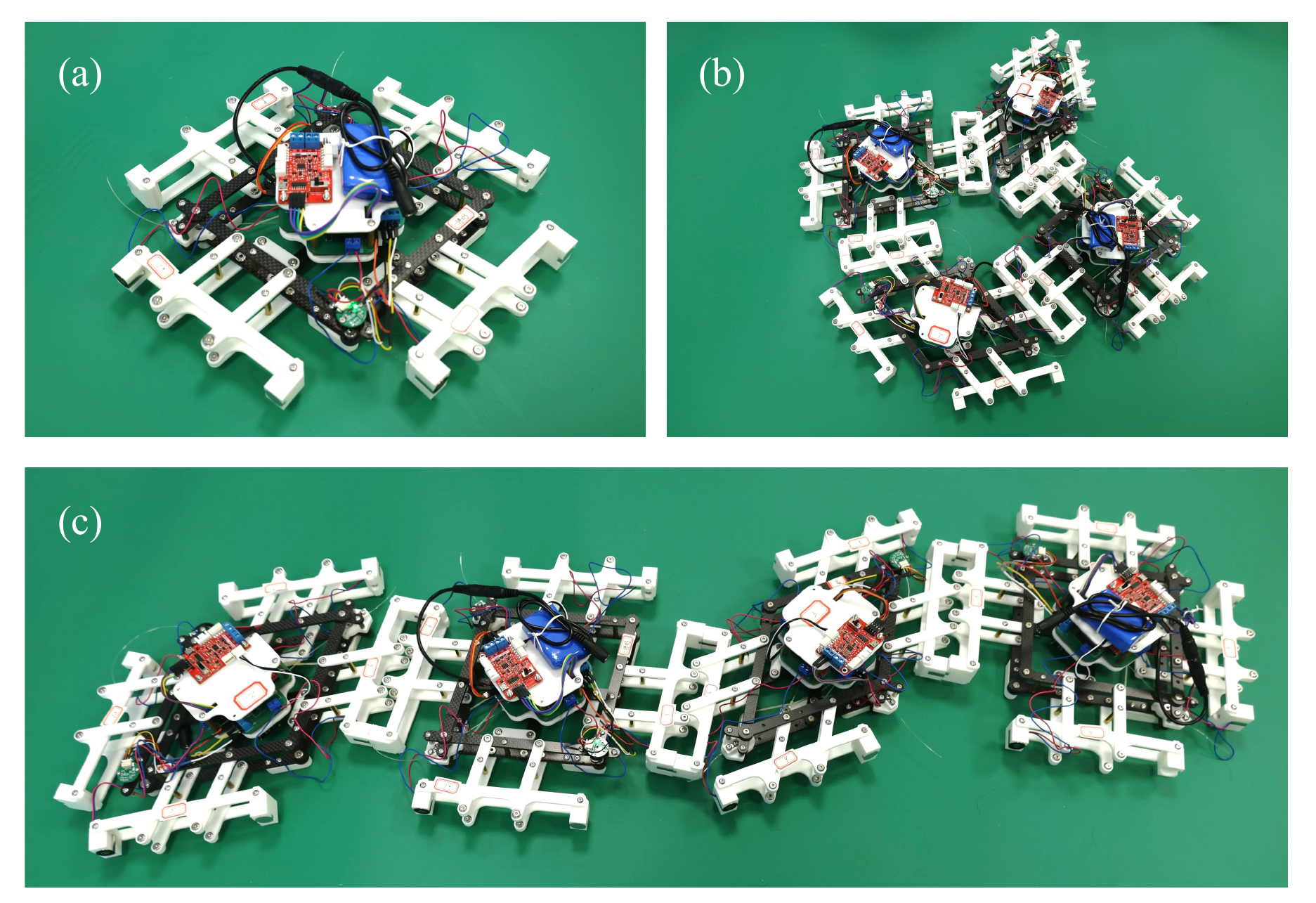}}
\caption{Prototype of RhoMorph. (a) Single Module. (b)  Lattice Assembly. (c) Chain Assembly.}
\label{figprototype}
\end{figure}

Our work primarily focuses on MSRRs with lattice attributes. For such robots, the most common reconfiguration motion is based on the disassembly of modules through the surrounding medium \cite{MARS,FTCP}. If the robotic system permits individual modules to detach, the system can, in theory, be entirely disassembled and reassembled to achieve any desired configuration. However, this approach faces a significant challenge: each module must precisely determine its designated position and successfully complete accurate docking. Another reconfiguration method is independent of the surrounding medium, relying instead on pivoting \cite{romanishin2013M-blocks,romanishin20153d-mblocks} or on the module’s intrinsic deformation capability to move toward neighboring voids \cite{pieber2018adaptive,piranda2021datom,qin2022trussbot}. The former approach imposes strict requirements on landing accuracy and does not allow modules to maintain intermediate states during transformation. In contrast, deformation-based approaches provide more stable connections and can maintain arbitrary intermediate configurations at any time.

In modular robots with deformable modules, modules are always interconnected through edges/faces in 2D/3D configurations \cite{gu2025modur,lyder2008odin,usevitch2020untethered}. During reconfiguration, the system ensures that every module remains connected to the overall structure through at least one edge or face, enabling stable and continuous reconfiguration motions. Furthermore, the relative coordinate frames defined within the system allow precise localization of individual modules, eliminating the need for external sensors to determine docking positions and thereby significantly improving docking reliability.
% 
% 强连通（可变形单元）的模块化机器人具有以下优点：
% 1.	模块间的运动不需要依靠外在的介质，所有的模块在每时每刻都是相互连通的，构成一个整体的，这非常适用于外太空作业。
% 2.	不再有游离的模块需要重新接入系统中，不再需要依靠感知进行对接，所有的对接具有统一的坐标系。
% MOPARAS（3D）	PARTS（2D）	M-blocks	Ours	

In this paper, we propose a two-dimensional deformable lattice MSRR named RhoMorph. Each module adopts a rhombus-shaped structure and incorporates a single DoF cable-driven actuation system at its center to enable expansion and contraction along its diagonal. This design provides both compactness and ease of control. Similar to other polyhedral MSRRs, RhoMorph modules can be tiled to form a rigid lattice structure when assembled. Alternatively, each module can be equivalently regarded as a revolute joint connected by rigid links, enabling the formation of serial chains that function as robotic manipulators. This dual nature allows RhoMorph to combine the advantages of both lattice-type and chain-type MSRRs. The prototype of RhoMorph shown as Fig.~\ref{figprototype}

This paper is organized as follows. Section II reviews related work on deformable MSRRs. Section III presents the mechanical design and electronic architecture of RhoMorph. Section IV analyzes the kinematics and introduces the principle of its morphpivoting motion. Section V validates RhoMorph’s stable, medium-free reconfiguration capability through a series of experiments.
%%%%%%%%%%%%%%%%%%%%%%%%%%%%%%%%%%%%%%%%%%%%%%%%%%%%%%%%%%%%%%%%%%%%%%%%%%%%%%%%
%%%%%%%%%%%%%%%%%%%%%%%%%%%%%%%%%%%%%%%%%%%%%%%%%%%%%%%%%%%%%%%%%%%%%%%%%%%%%%%%

% \section{Relate Work}
% \subsection{MSRRs with Lattice Attributes}
% MSRRs with lattice attributes一直是这个领域设计的热点方向，考虑到lattice的结构设计元素赋予msrr更好的刚性和重组能力。在\cite{liang2023decoding}的统计中有24/39个模块化机器人被划分到lattice的类型中。Roombot提出一种模块“互相帮助”来重构成各种构型，来辅助家具，并展现了很好的支撑能力和重构能力。\cite{空中模块化saldana2018modquad,水下模块化wang2019roboat}可利用水和空气介质集合在一起实现模块间的协同，获得一些个体不具备的能力。the M-TRAN is the first embody both the characteristics of chain and lattice type MSRR, 使得系统具备manipulation，可以通过两边的轮子实现重构。但是“互相帮助”的模式在重构时需要调用更多的模块，增加的控制的复杂度。其他几个则具有第一章中提及的问题。

% \subsection{Deformation MSRRs}
% \cite{pamecha2002useful}很早就提出了一种六杆结构的平面机器人系统，enables each module to move around another while remaining connected at all times during this motion，但是他在六个顶点处都安置了舵机，极大地增加了单个模块的控制难度，随着模块数量的上升，系统的控制难度急剧增加。\cite{gerbl2024parts}提出了一种三角形的可变形单元，每条边都可以伸缩以获取不同的三角形结构，但是它的构型中无法实现钝角三角形，所有的重构的动作只在理论上可行。

\section{Related Work}
\subsection{MSRRs with Lattice Attributes}
MSRRs with lattice characteristics have been a major focus of research, as lattice-style designs inherently provide high stiffness and facilitate efficient reconfiguration. According to the survey in \cite{liang2023decoding}, 24 out of 39 modular robotic systems can be classified as lattice-type MSRRs. Roombots \cite{sproewitz2009roombots} introduced a design that allows modules to assist each other during reconfiguration, demonstrating strong structural support and versatile reconfiguration capabilities. Similarly, ModQuad \cite{saldana2018modquad} and Roboat \cite{wang2019roboat} leverage air and water media, respectively, to achieve cooperative behaviors among modules, thereby achieving capabilities unattainable by individual units. The M-TRAN system \cite{kurokawa2008M-TRANIII} was the first to integrate features of both chain-type and lattice-type MSRRs, enabling manipulation capabilities and reconfiguration via wheels mounted on its edges. However, the assistance-based reconfiguration strategy typically requires a larger number of modules and leads to increased control complexity, while other medium-based lattice-type MSRRs exhibit the limitation discussed in Section~I.

\subsection{Deformation MSRRs}
Deformable MSRRs employ modules with variable geometry, enabling more versatile reconfiguration strategies compared with rigid lattice- or chain-type systems. A planar modular robotic system based on a six-bar linkage was introduced in \cite{pamecha2002useful}, allowing each module to move around another while maintaining connectivity throughout the motion. However, actuators were installed at all six vertices, resulting in a high control burden that scales rapidly with the number of modules. In \cite{gerbl2024parts}, a triangular deformable module was presented, with edges capable of extending and contracting to achieve various triangular configurations. Despite this theoretical versatility, the design cannot achieve highly obtuse triangles, which imposes certain kinematic constraints on reconfiguration. Overall, current deformation-based MSRRs highlight the potential of variable-geometry modules but also face challenges such as increased mechanical complexity, scalability issues, and limited demonstrated configurations.

\subsection{Comparison with previous MSRRs}
The RhoMorph integrates many functionalities commonly found in advanced lattice-type MSRRs, such as locomotion and inter-module connection/separation. In addition, it inherits certain features of chain-type MSRRs, such as manipulation. However, the primary focus of RhoMorph is achieving stable, medium-independent reconfiguration. A comparison with representative MSRR systems is presented in Table~I. Compared with Metamorphic robots and PARTS, RhoMorph leverages a single degree of freedom (DoF) to simplify reconfiguration control to the greatest extent, while offering a more stable and robust reconfiguration process than M-blocks.

\begin{table}[t]
\centering
\caption{Comparison with previous MSRRs}
\label{tab:msrr_comparison}
\begin{tabular}{lcccc}
\toprule
\textbf{MSRR System} & \textbf{DoF} & \textbf{Actuation} &\textbf{Approach} & \textbf{Ports} \\
\midrule
Metamorphic~\cite{pamecha2002useful} & 3 & 6 & Deformation & 6  \\
PARTS~\cite{gerbl2024parts}         & 3 & 3 & Deformation & 3  \\
M-Blocks~\cite{romanishin2013M-blocks} & 1 & 1 & Pivoting    & 6 \\
RhoMorph (Ours)                       & 1 & 1 & Deformation & 4 \\
\bottomrule
\end{tabular}
\end{table}

%%%%%%%%%%%%%%%%%%%%%%%%%%%%%%%%%%%%%%%%%%%%%%%%%%%%%%%%%%%%%%%%%%%%%%%%%%%%%%%%
%%%%%%%%%%%%%%%%%%%%%%%%%%%%%%%%%%%%%%%%%%%%%%%%%%%%%%%%%%%%%%%%%%%%%%%%%%%%%%%%
% section 3
\section{Mechanism Design and Electronics}

\begin{figure*}[t]
\centerline{\includegraphics[width=\linewidth]{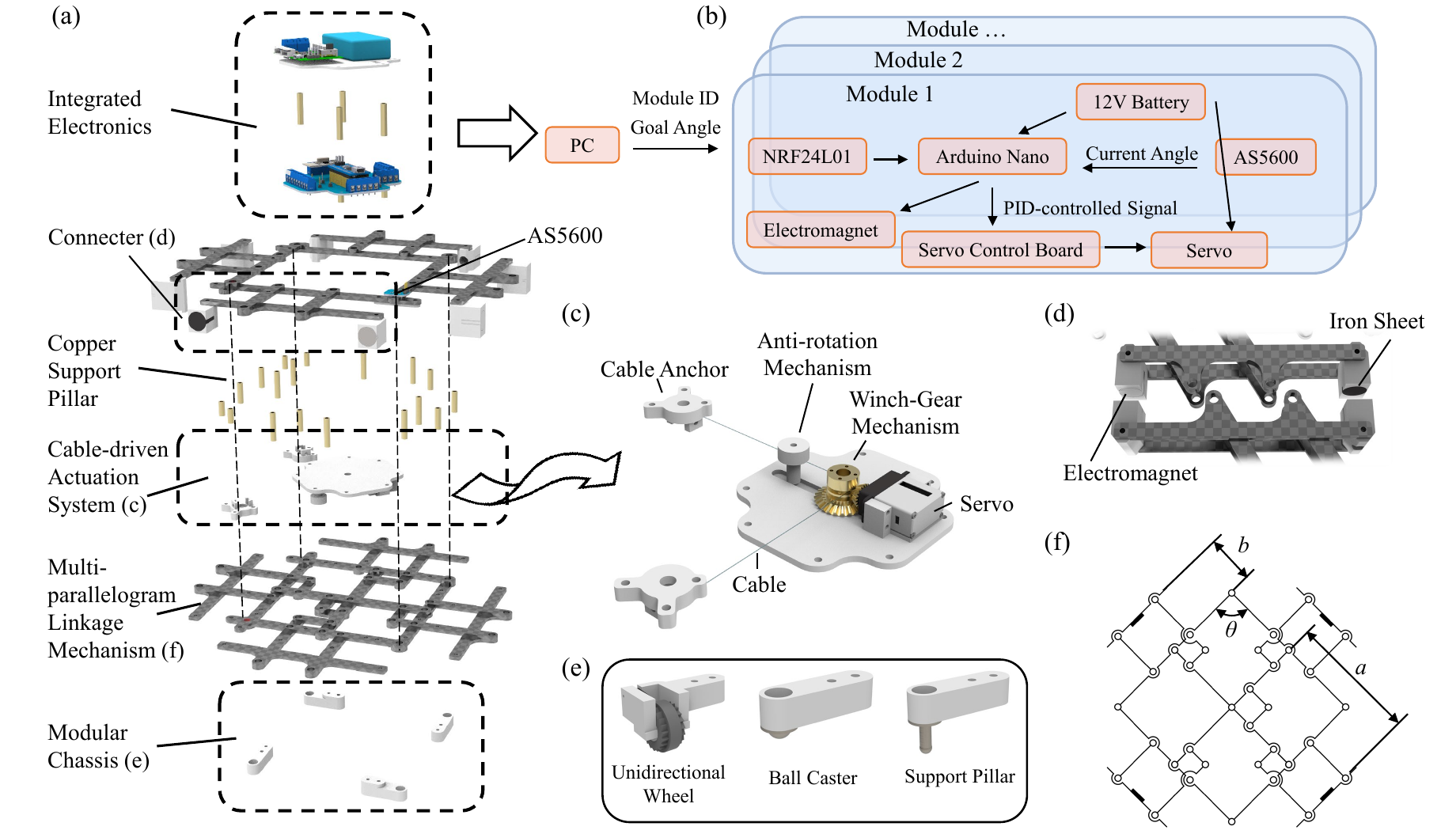}}
\caption{Mechanism Design and Electronics of the RhoMorph. (a) Exploded view of the RhoMorph’s components. (b) Electronic architecture layout. (c) Actuation system diagram. (d) Pair of hermaphroditic connectors. (e) Modular chassis design. (f) Schematic of the multi-parallelogram linkage mechanism.}
\label{figrhombot}
\end{figure*}
RhoMorph module is composed of multiple parallelogram linkages forming the skeleton, a cable-driven actuation system, four symmetrically distributed connectors, embedded electronics, and additional components necessary for full functionality. The exploded view is shown in Fig.~\ref{figrhombot}(a).

\subsection{Skeleton and Actuation System}
The skeleton, illustrated in Fig.~\ref{figrhombot}(f), consists of multi-parallelogram linkages, which provides the module’s fundamental morphing capability and serves as the foundation for integrating other components. With the exception of the central parallelogram structure, which forms a slot-like constraint with an anti-rotation mechanism to restrict diagonal movement and prevent undesired rotation about the center, the entire structure is centrally symmetric. The skeleton is fabricated from carbon fiber plates, providing high stiffness while preserving a compact overall form factor. The geometry is designed to eliminate component interference and ensure that the folding angle $\theta$ remains within the specified range, with half of the rhombus side length defined as $a$.
\begin{table}[t]
\caption{Parameters and Specifications of RhoMorph}
\begin{center}
\begin{tabular}{c p{5cm} c} % Adjusted column width for clarity
\toprule
\textbf{Item} & \textbf{Definition} & \textbf{Value} \\
\midrule
$a$   & Half of the rhombus side length & 140 mm \\
$b$   & Position of the electromagnet installation & 70 mm \\
$\theta$ & Folding angle & [45°, 135°] \\
$T$ & Output torque of the servo motor & 4.5 kg·cm \\
$Z_1$ & Number of teeth on the output gear & 12 \\
$Z_2$ & Number of teeth on the input gear & 24 \\
$r$ & Radius of the winch & 5 mm \\
$F_e$ & Standard holding force of the electromagnet & 25 N \\
\bottomrule
\end{tabular}
\label{tab1}
\end{center}
\end{table}

The actuation system employs a bevel gear transmission to connect the servo motor to the driving cable, shown as Fig.~\ref{figrhombot}(c). This design allows the servo to be mounted flat, minimizing the vertical dimension of the overall structure, and simultaneously increases actuation torque. The actuation torque $M_d$ generated between adjacent edges is calculated as:
\begin{equation}
    M_d = \frac{T \cdot Z_1}{r \cdot Z_2} \cdot 2a\sin\left(\frac{\theta}{2}\right)
    \label{eq:actuation_torque}
\end{equation}
where $T$ is the output torque of the servo, $Z_1$ and $Z_2$ are the numbers of teeth of the bevel gear pair, and $r$ is the radius of the winch.  

\subsection{Connector Design}
The connector consists of a normally-on electromagnet and a circular iron plate, shown as Fig.~\ref{figrhombot}(d). The electromagnet actively attracts the iron plate of another connector, creating a hermaphroditic connection. As a normally-on device, the electromagnet maintains its holding force without power, ensuring that modules remain securely connected even during power loss. When a module needs to detach from the robot system, it only deactivates its own electromagnets, while the mating connectors continue to provide resistance torque. The resisting torque is modeled as  
\begin{equation}
    M_f = F_e \cdot \left(2a-b\right) + \varepsilon
    \label{eq:resisting_torque}
\end{equation}
where $\varepsilon$ represents the additional frictional resistance at the physical contact surface. To enable single-sided disconnection, the design ensures that $M_f < M_d$. Experimental measurements indicate that $\varepsilon$ is approximately 0.84~kg$\cdot$cm. The complete structural design parameters are summarized in Table~II.

\subsection{Electronics}
The electronic architecture of the RhoMorph module is illustrated in Fig.~\ref{figrhombot}(b). The system integrates a wireless communication module (nRF24L01) for multi-module coordination, an Arduino Nano microcontroller, an AS5600 magnetic rotary encoder for measuring the folding angle $\theta$, electromagnets for inter-module connections, a servo motor, and a dedicated servo driver board. All electronic components are mounted on custom printed circuit boards (PCBs), resulting in a compact and modular electrical layout. Power is supplied by a 12~V lithium battery with step-down regulators providing the required voltages for each subsystem.

\subsection{Modular Chassis}
The chassis of the RhoMorph module adopts a modular design, shown as Fig.~\ref{figrhombot}(e), allowing it to be equipped with unidirectional wheels, ball casters, or support pillars. When equipped with support pillars, the module is anchored to the ground, enabling it to serve as a stable base for the entire system. Ball casters are the default choice for general deployment and are used as the base in all modules presented in this work, providing low-friction omnidirectional support. In contrast, unidirectional wheels enable individual modules to achieve independent mobility but are not considered in the design or experiments of this paper.

%%%%%%%%%%%%%%%%%%%%%%%%%%%%%%%%%%%%%%%%%%%%%%%%%%%%%%%%%%%%%%%%%%%%%%%%
\section{Motion Analysis}
\subsection{Kinematic Modeling}
Since the kinematic system of modular robots varies with changes in the number of modules or their topological arrangement, deriving a single, invariant kinematic model is impractical. A module-based kinematic modeling approach is therefore widely adopted\cite{tu2025locomotion}, where each kinematic module is denoted as $M_i$($i \in \mathbb{N}$) and connected according to topology, resulting in a kinematic tree.

The RhoMorph module is modeled as a two-dimensional quadrilateral. Among its four edges, $E_0$ is connected either to the parent node or to the base, while the remaining three edges, labeled $E_1$, $E_2$, and $E_3$ in counterclockwise order, serve as interfaces for connecting to other modules. For a given module $M_i$, a reference coordinate frame $\{O_i\}$ is defined at the midpoint of its edge $E_{i0}$. The $y_i$-axis is oriented toward the geometric center of the module, the $z_i$-axis is normal to the plane and directed outward, and the $x_i$-axis is determined according to the right-hand rule.

\begin{figure}[t]
\centerline{\includegraphics[width=\linewidth]{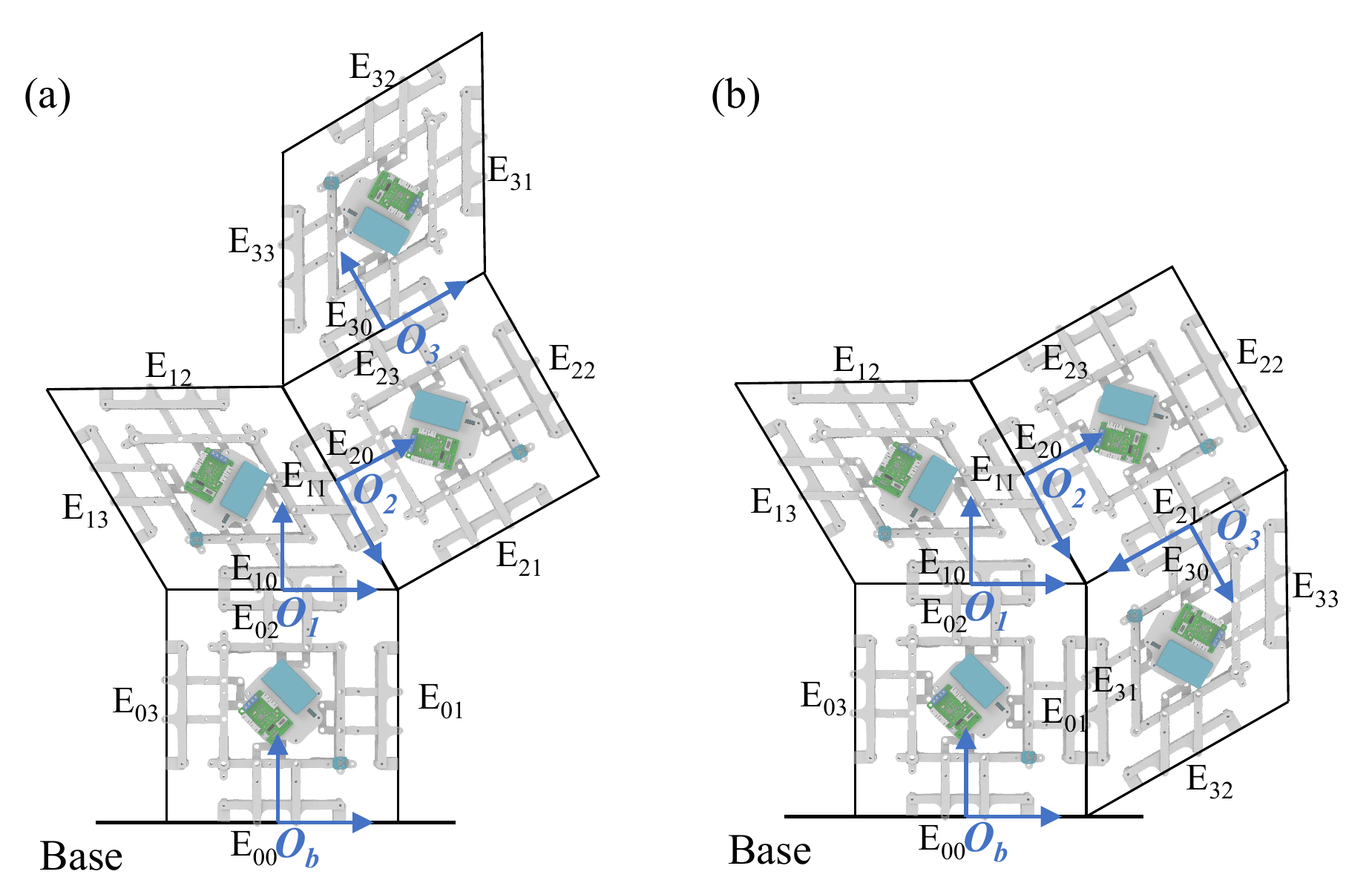}}
\caption{Two kinematics of RhoMorph with configuration representations and reference coordinate frames indicated. (a) Serial connection of four modules. (b) Parallel connection of four modules.}
\label{figkinemics}
\end{figure}

The transformations from the midpoint of $E_{i0}$ to the midpoints of $E_{i1}$, $E_{i2}$, and $E_{i3}$ are expressed as 
\begin{equation}
{}^{0}_{1}T_i =
\begin{bmatrix}
R_z(\sigma_i+\pi) & 
\begin{bmatrix}
a+a\cos\sigma_i \\ a\sin\sigma_i \\ 0
\end{bmatrix} \\
0 \ 0 \ 0 & 1
\end{bmatrix}
\label{eq:T01}
\end{equation}

\begin{equation}
{}^{0}_{2}T_i =
\begin{bmatrix}
I_{3} & 
\begin{bmatrix}
2a\cos\sigma_i \\ 2a\sin\sigma_i \\ 0
\end{bmatrix} \\
0 \ 0 \ 0 & 1
\end{bmatrix}
\label{eq:T02}
\end{equation}

\begin{equation}
{}^{0}_{3}T_i =
\begin{bmatrix}
R_z(\sigma_i) & 
\begin{bmatrix}
a-a\cos\sigma_i \\ a\sin\sigma_i \\ 0
\end{bmatrix} \\
0 \ 0 \ 0 & 1
\end{bmatrix}
\label{eq:T03}
\end{equation}

where $R_z(\cdot)$ represents a rotation matrix about the $z$-axis. The parameter $\sigma_i$ characterizes the folding--unfolding degree of freedom of module $M_i$, corresponding to the interior angle between edges $E_0$ and $E_3$, which can be mapped from $\theta_i$ based on the connection configuration. Initially, $\sigma_i = \theta_i$; however, after the modules undergo reconfiguration, the reference edge $E_0$ is redefined, requiring a remapping of $\sigma_i$.

\begin{algorithm}[t]
\caption{Reconfiguration Strategy}
\tcc*{All operations are performed on the module system $Ktree$}
\KwIn{$Con\_DisconPair$}

InitializeKTree() \;
\While{$Con\_DisconPair$ is not empty}{
    $NewCon,NewDiscon \gets Con\_DisconPair.\mathrm{Pop}()$ \; 
    $Edges \gets$ IdentifyEdges$(NewCon)$ \; 
    Morphpivoting($edges$, $NewCon$, $NewDiscon$) \;
}
\hrule
\textbf{function} Morphpivoting($Edges$, $NewCon$, $NewDiscon$) \;
\Indp 
Connect($Edges$, $NewCon$) \;
Disconnect($NewDiscon$) \;
$Child \gets$ DetermineChild$(NewDiscon)$ \;
AssignNewParent($Child$) \;
\Indm
\textbf{end} \\[3pt]
\hrule
\textbf{function} AssignNewParent($M$) \;
\Indp 
\ForEach{$M_{near}, M_{near}\_edge$ in $M.conns$}{
    $Status, Path \gets $IsConn$(Root, M_{near})$ \;
    \If{$Status$ \textbf{and} $M \notin Path$}{    
        Connect($M$, $M_{near}$, $M_{near}\_edge$) \;
        UpdateInform($M$) \;
        RebuildConns() \;
        \textbf{break} \;
    }
}
\Indm
\textbf{end}
\end{algorithm}

When constructing the kinematic tree, one module is always selected as the root module, with its edge $E_0$ fixed on the base. The coordinate frame $\{O_0\}$ of the root module is simultaneously defined as the base frame of the entire robotic system, denoted as $\{O_b\}$. 
If a closed-loop structure exists, it can be converted into a tree structure by disconnecting one of the topological connections and introducing additional constraints. Accordingly, the forward kinematics, which is the transformation from the base frame $\{O_b\}$ to the 
end-module frame $\{O_e\}$, can be expressed as:
\begin{equation}
    {}^{b}_{e}T = {}^{0}_{k_0}T_0 \,{}^{0}_{k_1}T_1 \,{}^{0}_{k_2}T_2\, \cdots \,{}^{0}_{k_e}T_e
\end{equation}
where $K=\{k_0,k_1,k_2,\ldots,k_e\}$ denotes the sequence of selected interfaces.

Fig.~\ref{figkinemics} illustrates the kinematic trees with serial and parallel topologies. 
In Fig.~\ref{figkinemics}(a), the pose of any module edge can be obtained from (4). 
In Fig.~\ref{figkinemics}(b), disconnecting $M_0$–$M_3$ gives the serial path $M_0M_1M_2M_3$, 
while disconnecting $M_2$–$M_3$ gives the path $M_0M_3$. 
Both yield the same center position of $M_3$, leading to the basic constraint:

\begin{equation}
    {}^{0}_{2}T_0 \,{}^{0}_{1}T_1 \,{}^{0}_{1}T_2 \,{}^{0}_{c}T_3 
    = {}^{0}_{1}T_0 \,{}^{0}_{c}T_3'
\end{equation}

where ${}^{0}_{c}T_3$ is the transformation from $E_0$ to the module center, 
with ${}^{0}_{c}T_3\,{}^{0}_{c}T_3 = {}^{0}_{2}T_3$. The ${}^{0}_{c}T_3$ on both sides of the equation differ because $\sigma$ is defined differently, and the prime ${}^{\prime}$ distinguishes them.

\subsection{Reconfiguration Strategy}
\begin{figure}[t]
\centerline{\includegraphics[width=\linewidth]{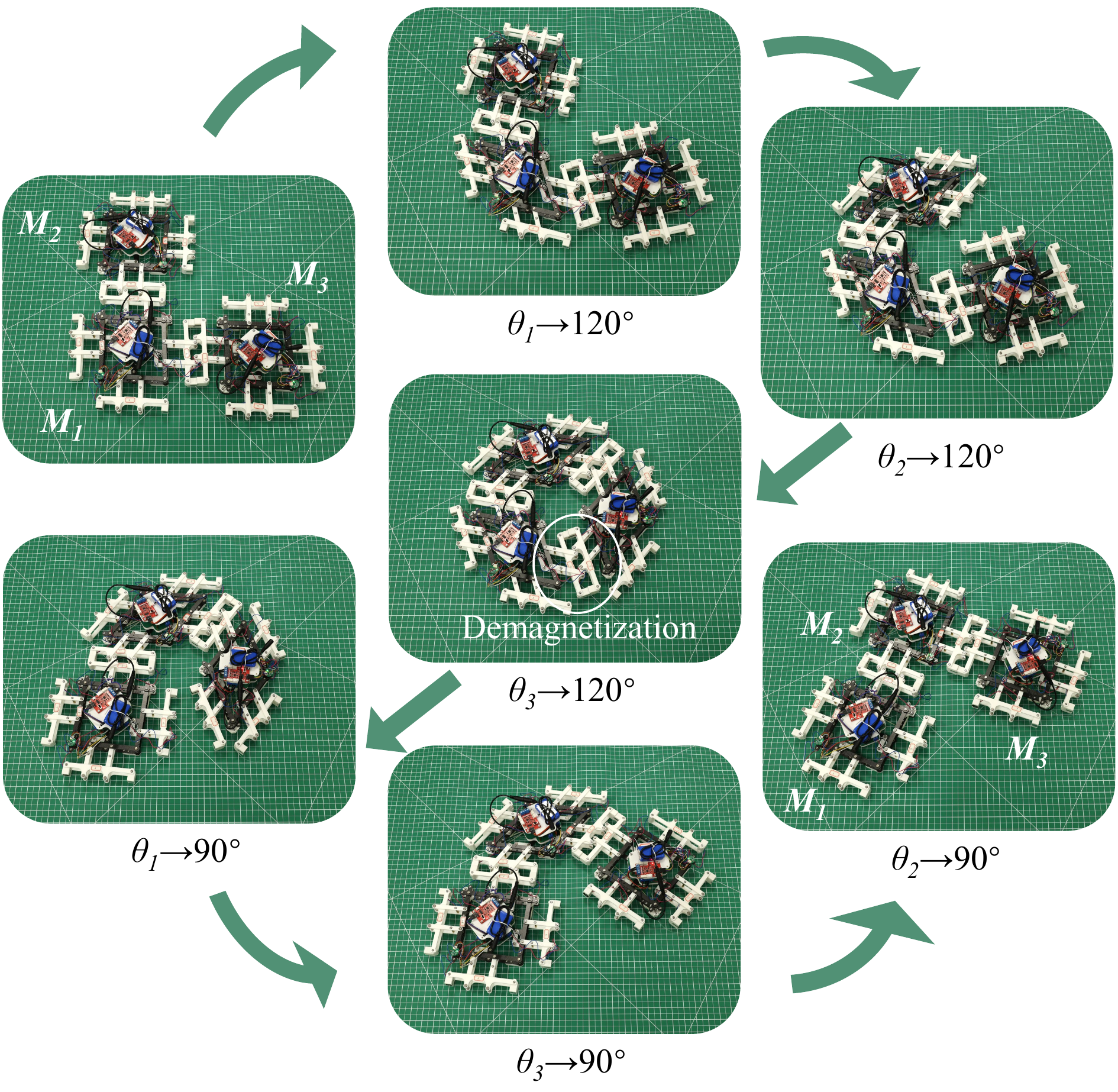}}
\caption{Demonstration of a morphpivoting operation with three RhoMorphs.}
\label{fig:threeprototype}
\end{figure}

Based on the above kinematics, we can construct an initial kinematic model for any configuration of the RhoMorph and determine the position of each module. Building on this model, we focus on the motion capabilities of the RhoMorph. The RhoMorph supports most functions of modular self-reconfigurable robots (MSRRs), such as morphing, locomotion, attachment, and detachment. However, these fundamental motions primarily serve as the basis for reconfiguration, which we define as incrementally transitioning the modules from one topological connection state to another. In this work, we introduce morphpivoting as a novel primitive motion that acts as the elemental operation enabling such reconfiguration. Each morphpivoting process consists of four sequential steps: morphing, connection, disconnection, and morphing.

Algorithm 1 presents a systematic strategy for executing multiple morphpivoting operations in an orderly manner, based on specified connection and disconnection pairs. First, a topological tree is constructed to represent the modular configuration, initialized according to the geometric positions of the modules and their connectivity relationships (Line~1). For each specified connection pair, the algorithm determines the corresponding module edges involved in the operation and invokes the morphpivoting motion procedure (Lines~2--7). This procedure establishes the required new connection, removes the specified disconnection, and updates the topological tree, with a key focus on inferring parent--child relationships among the affected modules and reassigning parent nodes to ensure a consistent hierarchical structure (Lines~8--13). Subsequently, the \textsc{AssignNewParent} function identifies a valid neighboring module to serve as the new parent for the child module, preserving the overall tree connectivity (Lines~14--24). The child module is reattached to this parent through the appropriate edge, and the labeling and connectivity information of all affected modules are updated accordingly. Finally, the \textsc{UpdateInform} function recalculates $E_0$ and remaps $\sigma$, which modifies the kinematic tree structure and necessitates its regeneration after each morphpivoting operation, as specified in (6).

%%%%%%%%%%%%%%%%%%%%%%%%%%%%%%%%%%%%%%%%%%%%%%%%%%%%%%%%%%%%%%%%%%%%%%%%
\section{Experiments}
\subsection{Reconfiguration}

\begin{figure}[t]
\centerline{\includegraphics[width=\linewidth]{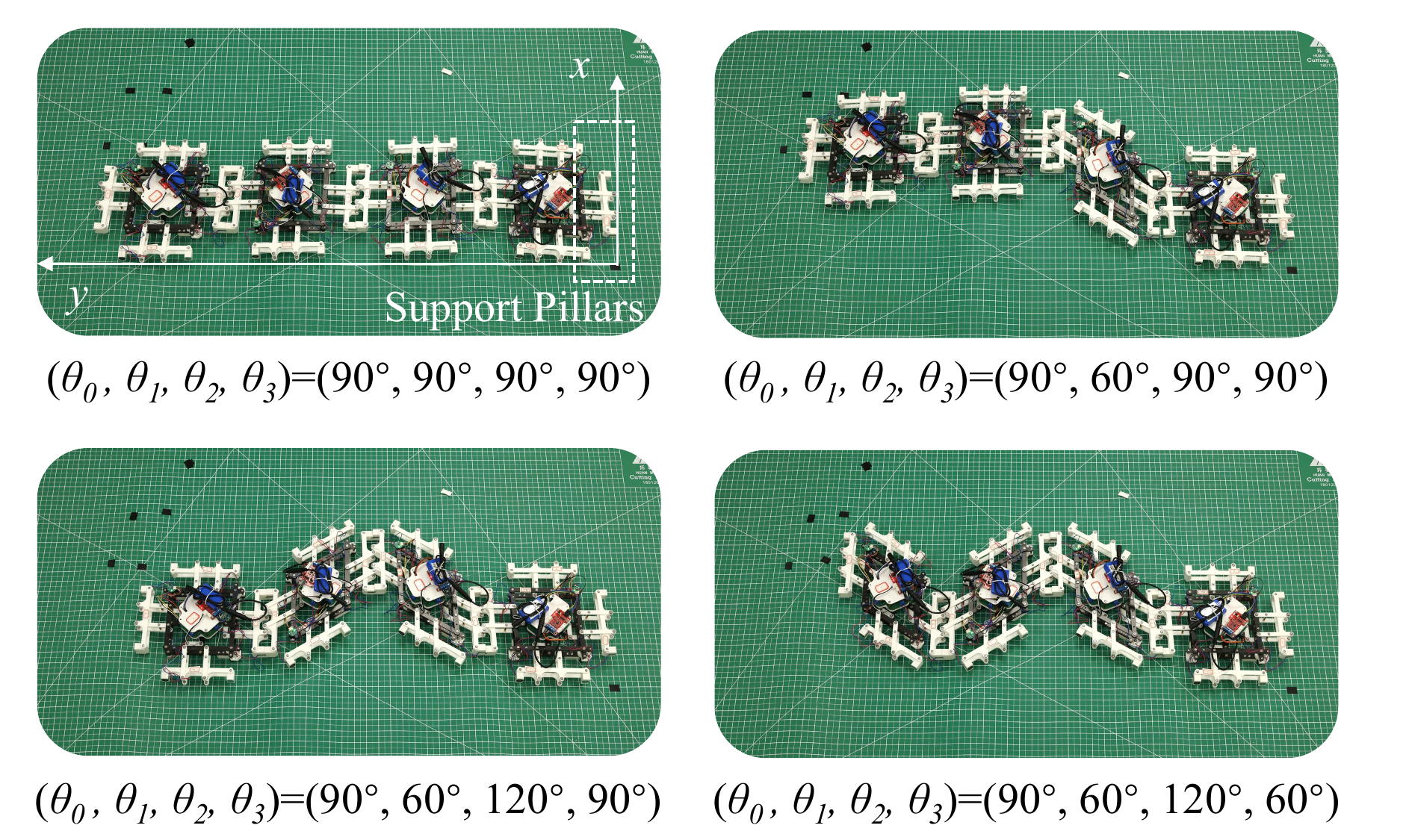}}
\caption{Kinematic Accuracy of RhoMorph's Chain Form.}
\label{figshiyan}
\end{figure}

\begin{figure}[t]
\centerline{\includegraphics[width=\linewidth]{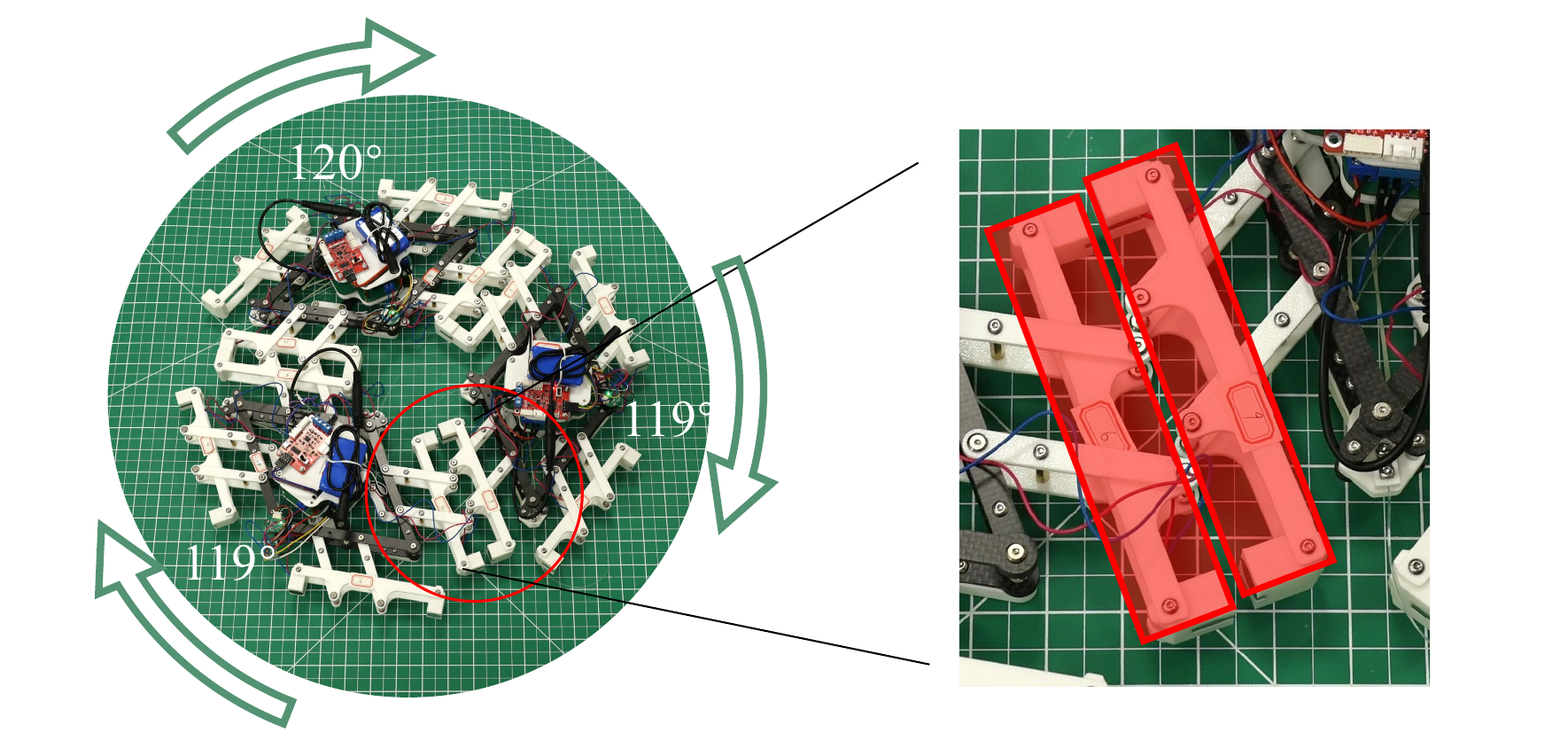}}
\caption{Docking alignment after multiple morphpivotings.}
\label{figdocking}
\end{figure}

\begin{figure*}[t]
\centerline{\includegraphics[width=\linewidth]{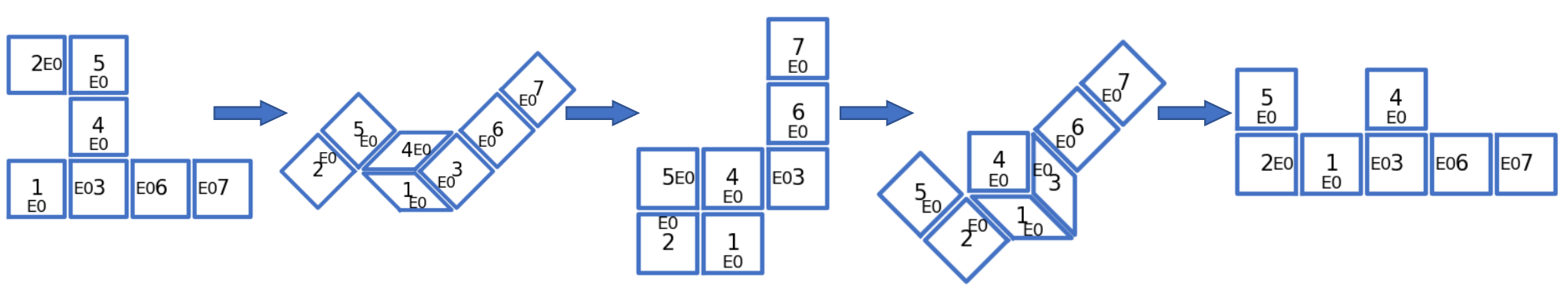}}
\caption{Simulation of reconfiguration using morphpivoting, transforming the structure from a horizontally placed “$\mu$” shape to a horizontally placed “F” shape.}
\label{figreconfig}
\end{figure*}

Fig.~\ref{fig:threeprototype} illustrates a morphpivoting operation, which consists of morphing, connection, separation, and subsequent morphing between RhoMorphs. Three modules, denoted as $M_1$, $M_2$, and $M_3$, are initially configured with $\theta = 90^\circ$. Their angles are then sequentially adjusted to $\theta = 120^\circ$ to form a triangular loop. A new connection is established between $M_2$ and $M_3$, while the connection between $M_1$ and $M_3$ is released by demagnetizing the electromagnet. Finally, the angles are sequentially restored to $\theta = 90^\circ$.

Due to the limited number of fabricated modules, simulations were conducted to demonstrate the reconfiguration strategy described in Section~IV. The transformation process is shown in Fig.~\ref{figreconfig}, where seven modules arranged in a horizontally placed “$\mu$” configuration are reconfigured into a horizontally placed “F” through multiple morphpivoting operations. The edge indices before and after the transformation are labeled to illustrate the redefinition of edges as the system’s topology changes.

\subsection{Kinematic Accuracy Evaluation}
In a chain configuration, the system is expected to perform manipulator-like operations, since each module can be equivalently modeled as a revolute joint connected by rigid links. The kinematic analysis of the end-effector position was presented in Section~IV, and here we experimentally validate the accuracy of the end position on the physical prototype. The experiment was conducted on a green fabric with printed scales to capture the position of each module, with four experimental snapshots shown in Fig.~\ref{figshiyan}. During the experiment, the $\theta$ values of the modules were adjusted to induce swinging motions of the chain structure. For each configuration, the intersection point of $E_1$ and $E_2$ on the end module was measured as the ground-truth position, while the theoretical position was obtained through kinematic analysis. A comparison between the measured and calculated positions shows an RMSE of approximately 4.77~mm along the $x$-axis and 14.96~mm along the $y$-axis. The larger error in the $y$-direction is attributed to longitudinal friction that hindered extension along this axis, despite the use of ball casters.

\subsection{Docking Accuracy Evaluation}
To validate the stability of the reconfiguration process, the procedure illustrated in Fig.~\ref{fig:threeprototype} was repeated for seven consecutive trials. In the final morphpivoting step, the $\theta$ angles of modules $M_1$ and $M_3$ were set to 119$^{\circ}$ to bring them into close proximity for docking observation, as shown in Fig.~\ref{figdocking}. The red-highlighted region indicates the relative offset between the two connectors, which remains within the acceptable tolerance.

\section{Discussion}
\begin{figure}[t]
\centerline{\includegraphics[width=\linewidth]{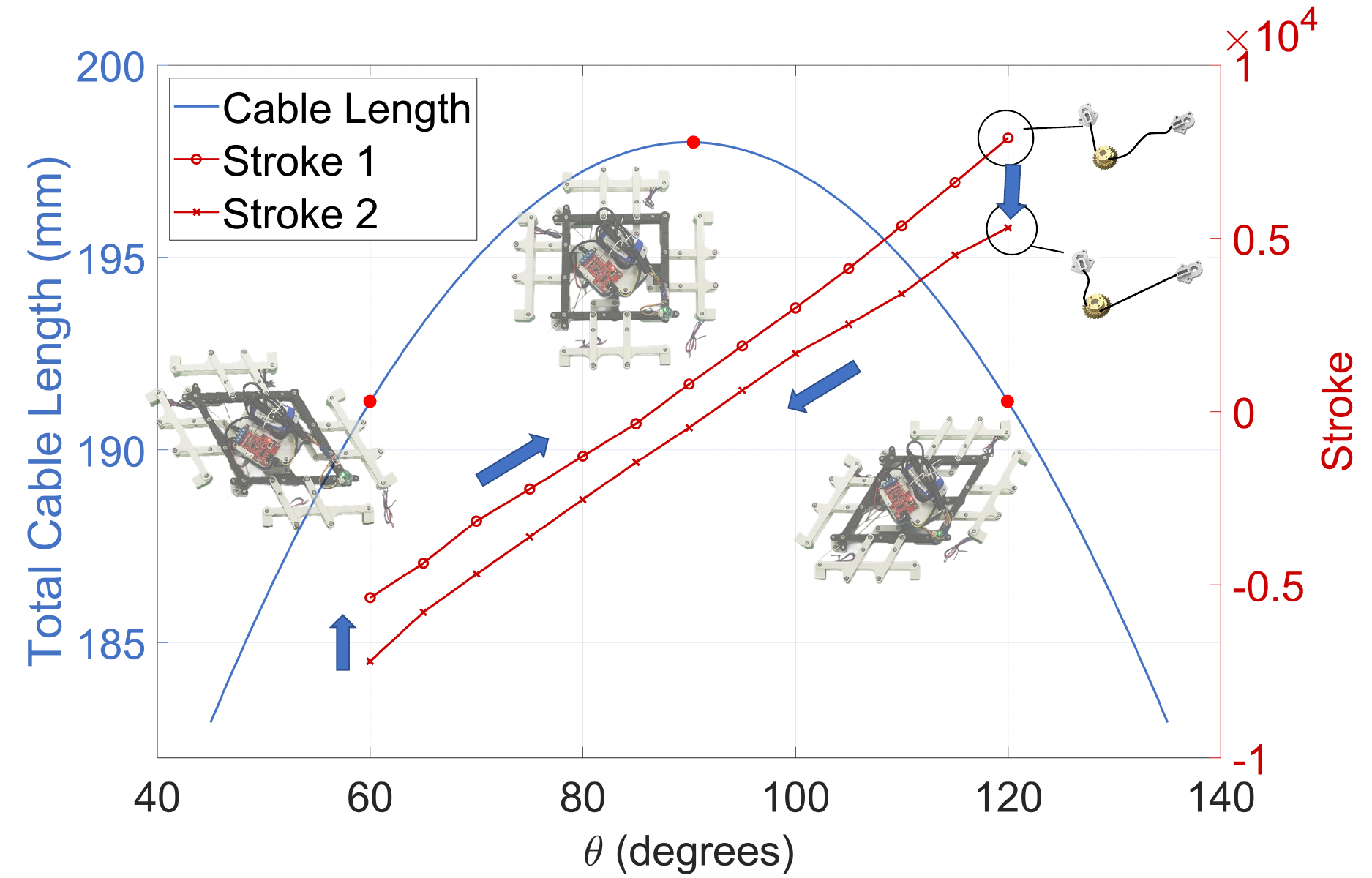}}
\caption{Total cable length and servo strokes in forward and reverse directions plotted against the folding angle $\theta$.}

\label{figcable}
\end{figure}

In practical implementation, imperfect cable tensioning and variations in the diagonal length during the morphing process result in a slight offset between the servo strokes in the forward (increasing $\theta$) and reverse (decreasing $\theta$) rotations, as illustrated in Fig.~\ref{figcable}. This offset corresponds to approximately 1500 encoder counts, or about $131^\circ$ given a resolution of 4095, and is unavoidable near the end ranges of motion. In the physical prototype, two springs were incorporated to mitigate cable slack, ensuring that this issue does not fundamentally affect system performance. For future improvements, an irregular winch design could be considered to further reduce this effect.

%%%%%%%%%%%%%%%%%%%%%%%%%%%%%%%%%%%%%%%%%%%%%%%%%%%%%%%%%%%%%%%%%%%%%%%%
\section{Conclusion}
This paper presents a novel lattice-based MSRR, RhoMorph, which demonstrates stable and medium-independent reconfiguration capabilities. RhoMorph is driven by a single servo, offering high controllability for individual modules while enabling advanced MSRR functionalities such as morphing, connection/disconnection, and more. Its lattice structure allows for planar tiling, achieving the high rigidity inherent in lattice-based MSRRs. At the same time, simplifying each module to a rotational pair and linkage provides the flexibility characteristic of chain-based MSRRs. The reconfiguration capabilities and docking precision of the RhoMorph system were experimentally validated.

Several avenues for future research exist. For instance, the modular chassis could be replaced with unidirectional wheel, to investigate locomotion for individual modules and synchronized movement for multi-module, such as snake-like configurations. In terms of reconfiguration planning, a critical next step is the development of an autonomous algorithm that uses the morphpivoting motion as a primitive. Such an algorithm would generate optimal action sequences, overcoming the current limitation of manually specified connection/disconnection pairs. This will be fundamental to achieving large-scale and fully autonomous applications for the RhoMorph system.

%                                   % the textheight of the last page by a suitable amount.
%                                   % This command does not take effect until the next page
%                                   % so it should come on the page before the last. Make
%                                   % sure that you do not shorten the textheight too much.

% \let\balance\relax

\bibliographystyle{IEEEtran}
\bibliography{references}

\end{document}